\documentclass[10pt]{article}
\usepackage[T1]{fontenc}
\usepackage{graphicx,xcolor}
\usepackage{amssymb}
\usepackage{amsmath}

\def\x{\mathbf{x}}

\begin{document}

\title{\large Parameter Tuning of the Firefly Algorithm by Standard Monte Carlo and Quasi-Monte Carlo Methods}

\author{Geethu Joy$^{1,2}$
\thanks{Corresponding author, email:  gj219@live.mdx.ac.uk},
 Christian Huyck$^{1}$, Xin-She Yang$^{1}$ }

\date{\small 1. School of Science and Technology, Middlesex University, London NW4 4BT, UK \\
2. Computer Engineering and Informatics, Middlesex University Dubai, \\ Dubai Knowledge Park, P. O. Box 500697, Dubai, United Arab Emirates. }

\maketitle

\begin{abstract}
  Almost all optimization algorithms have algorithm-dependent parameters, and the setting of such parameter values can significantly influence the behavior of the algorithm under consideration. Thus, proper parameter tuning should be carried out to ensure that the algorithm used for optimization performs well and is sufficiently robust for solving different types of optimization problems. In this study, the Firefly Algorithm (FA) is used to evaluate the influence of its parameter values on its efficiency. Parameter values are randomly initialized using both the standard Monte Carlo method and the Quasi Monte-Carlo method. The values are then used for tuning the FA.
  Two benchmark functions and a spring design problem are used to test the robustness of the tuned FA. From the preliminary findings, it can be deduced that both the Monte Carlo method and Quasi-Monte Carlo method produce similar results in terms of optimal fitness values. Numerical experiments using the two different methods on both benchmark functions and the spring design problem showed no major variations in the final fitness values, irrespective of the different sample values selected during the simulations. This insensitivity indicates the robustness of the FA.

\end{abstract}

\noindent {\color{blue}{\bf Citation Details:} \\[7pt]
Geethu Joy, Christian Huyck, Xin-She Yang, Parameter Tuning of the Firefly Algorithm by Standard Monte Carlo and Quasi-Monte Carlo methods. In:
Franco, L., de Mulatier, C., Paszynski, M., Krzhizhanovskaya, V.V., Dongarra, J.J., Sloot, P.M.A. (eds) Computational Science – ICCS 2024. ICCS 2024. Lecture Notes in Computer Science, vol 14836. pp/ 242--253, (2024). 
Springer, Cham.
https://doi.org/10.1007/978-3-031-63775-9\_17
}

\section{Introduction}
Many problems in engineering design and industry can be formulated as
optimization problems with a main design objective, subject to multiple nonlinear constraints. Best design options correspond to the optimal solutions to such design optimization problems. To find such optimal solutions requires the use of sophisticated optimization algorithms and techniques~\cite{ref_article1,ref_book7}. A recent trend is to use nature-inspired algorithms to solve engineering design optimization problems because nature-inspired algorithms tend to be effective, flexible and easy to implement.

Almost all optimization algorithms and techniques, including nature-inspired algorithms,  have algorithm-dependent parameters, and these parameters need to be properly tuned. Tuning algorithm-specific parameters play a crucial role in determining the effectiveness of an algorithm, and the way these parameters are configured can significantly influence the performance of the algorithm under consideration~\cite{ref_article8,ref_book4}. Consequently, fine-tuning algorithmic parameters is a vital aspect of implementing and applying algorithms for solving problems in real-world scenarios~\cite{ref_book2,FA-Osaba-paper,FA-robot-paper}.

Ideally, an efficient method should be used for tuning or setting the parameters for a given algorithm so that the algorithm can obtain better results. However, such methods are not yet available, and thus tuning is largely empirical or experience-based. Therefore, tuning parameters for a given algorithm can still be a challenging task, especially for tuning nature-inspired optimization algorithms~\cite{ref_article2}. In addition,  parameter tuning can be problem-specific, and even with advanced tuning tools, a finely-tuned algorithm for one problem may not generalize well to other problems, leading to the need for re-tuning for each new problem or problem type and thus making it a time-consuming task in optimization.

In this study, two different methods, namely, the standard Monte Carlo (MC) and Quasi-Monte Carlo (QMC) methods, will be used to tune the parameters of the Firefly Algorithm (FA). The performance of the tuned FA will then be evaluated by comparing the fitness values obtained by FA using both MC and QMC. These numerical experiments may give some insights into the tuning efficiency of MC and QMC by investigating both the fitness values of problems with two different tuning methods and over different optimization problems.

\section{Literature Review of Parameter Tuning}

The literature of parameter tuning for evolutionary algorithms and metaheuristics is expanding, especially in the context of tuning parameters of nature-inspired metaheuristic algorithms. Here, a brief review is carried out on different approaches to parameter tuning, including online and offline approaches~\cite{ref_book7}.

From the perspective of parameter tuning, for a given algorithm,  a tuning tool (or a tuner) should be used to tune the algorithm first, and then use the tuned algorithm to solve a set of problems. Thus, there are three key components here: an algorithm, a tuner, and a problem set. Since these three components are involved simultaneously in tuning, it is possible that the parameter values tuned may depend on both the algorithm under consideration and the problems to be solved. Therefore, parameter settings can be algorithm-specific as well as problem-specific.

For a given algorithm, its parameters can be tuned first before it is used for solving optimization problems. This approach is usually called offline tuning. Other studies also indicate that it may be advantageous to vary parameters during iterations, and this approach is often referred to as online tuning~\cite{ref_article8,ref_book2}. Parameter tuning can be carried out either in sequence or in parallel and these different methods can be loosely divided into ten different categories:

\begin{itemize}
	\item Manual or brute force method
	\item Tuning by systematic scanning
	\item Empirical tuning as parametric studies
	\item Monte Carlo based method
	\item Tuning by design of experiments (DOE)
	\item Machine learning based methods
	\item Adaptive and automation methods
    \item Self-tuning method~\cite{ref_article9}
    \item Heuristic tuning with parameter control
    \item Other tuning methods
\end{itemize}	

Other parameter tuning methods include sequential optimization approaches, multi-objective optimization approaches, self-parameterization,
fuzzy methods as well as dynamic parameter adaptation approaches and hyper-parameter optimization~\cite{ref_article10,ref_article11}. Although extensive studies have been dedicated to exploring parameter tuning methods, both offline and online, a lack of comprehensive understanding persists regarding these methods. Sometimes, these methods may not perform as well as expected, and the reasons behind these unexpected outcomes remain elusive, underscoring the need for deeper insights. Some key issues in parameter tuning include~\cite{ref_book4}

\begin{enumerate}

\item {\it Non-universality}. It is not clear whether tuned parameters are inherently problem-specific and algorithm-specific, which may limit their generalization to different problem sets.

\item {\it High computational efforts}. Parameter tuning tends to be a computationally intensive task.  This poses a significant barrier to effective parameter tuning, necessitating the development of methods to minimize computational efforts.

\item {\it Lack of theoretical insights}. Despite a diverse spectrum of tuning methods available in the current literature, most rely on heuristic approaches without theory-based guidelines, lacking a clear understanding of their mechanisms and optimal conditions for applications. 	
\end{enumerate}

There are still some open problems related to parameter tuning. For example, it is not clear how to tune parameters in the most effective way and how mathematical theories can be applied to parameter tuning. In addition,  the practical implications of well-tuned parameters on algorithm convergence remain another open problem, thus highlighting the need for further research in this area.

\section{Tuning Parameters by MC and QMC}

Before the details of tuning parameters using the MC and QMC methods are discussed, the main idea of FA and its parameters are outlined.

\subsection{Firefly Algorithm}

The Firefly Algorithm (FA) is a nature-inspired algorithm that was developed by Xin-She Yang in 2008, based on the flashing characteristics and flying patterns of tropical fireflies~\cite{FA_paper}. FA has been applied to a diverse range of applications, including multi-modal optimization, multi-objective optimization~\cite{FA_paper,ref_book6}, clustering~\cite{FA-cluster-paper}, software testing~\cite{ref_article3},
vehicle routing problems~\cite{FA-Osaba-paper}, multi-robot swarming~\cite{FA-robot-paper} and others.
	
For a given optimization problem, its solution vector $\x$ is encoded
as the locations of fireflies. Thus, the locations of two fireflies
$i$ and $j$ correspond to two solution vectors $\x_i$ and $\x_j$, respectively.  The main updating equation of firefly locations or solution vectors is

\begin{equation}
		\x_{i}^{t+1} = \x_{i}^{t} + \beta e^{-\gamma r_{ij}^2}
	(\x_{j} - \x_{i}) + \alpha \epsilon_{i}^{t}, \label{Position-Update}
	\end{equation}
where the random number vector $\epsilon_i^t$ is drawn from a Gaussian normal distribution.  In addition, the distance $r$ between two solutions is given by the Euclidean distance or $L_2$-norm
\begin{equation}
\quad r_{ij} = \lVert x_{i}^{t} - x_{j}^{t} \rVert.
\end{equation}
The parameters to be tuned are the attractiveness parameter $\beta$,
the scaling parameter $\gamma$ and the randomization strength parameter $\alpha$. In most FA implementations, parameter $\alpha$ is further rewritten as
\begin{equation}
\alpha=\alpha_0 \theta^t,
\end{equation}
where $\alpha_0$ is its initial value, which can be set to $\alpha_0=1$. Here, $t$ is the pseudo-time or iteration counter, and $0<\theta<1$ is the parameter to be tuned, instead of $\alpha$.

\subsection{Monte Carlo  Method}

One approach to offline tuning is to use MC-based methods. In this study,
all the parameters in the FA are initialized randomly using MC and pseudo-random numbers that are uniformly distributed.   Pseudo-random numbers are random numbers generated using generators, which are used in computer programs. They are not truly random numbers and are generated in a deterministic way with some sophisticated permutations.

In essence, the MC method is a statistical sampling method with
statistical foundations and its errors tend to decrease as $O(1/\sqrt{N})$
where $N$ is the number of samples~\cite{ref_book1}. Though this inverse-square convergence
may be slow, it can work well in practice~\cite{ref_book4}, in comparison with manual or brute force methods.

In the current simulation for parameter tuning, the parameters of the FA are randomly initialized by drawing random samples from uniform distributions in a specific range of parameter values. Then, the discrete random samples are used as the parameter setting of the FA. With
such settings, the FA is executed to solve the given optimization problems, such as the benchmark functions and the spring design problem~\cite{ref_article1,ref_book3}.

\subsection{Quasi-Monte Carlo Method}

To obtain better estimates, the standard MC method requires a large number of samples. Theoretical analysis and studies from various applications suggest that a quasi-Monte Carlo method
can potentially speed up the convergence because
its errors decrease as $O(1/N)$ under certain conditions. Such QMC methods use low-discrepancy sequences or quasi-random numbers, and such sequences require some careful generation and random scrambling of the initial  sequences~\cite{ref_article4,ref_book1,ref_book3}.
Therefore, this study also uses QMC to tune parameters in the FA and comparison with the standard MC will be carried out.

For the generation of quasi-random numbers, there are efficient algorithms
such as van der Corput sequence, Sobol sequence and Halton sequence.
Most of these sequences will generate quasi-random numbers in the
interval between $0$ and $1$. In the current simulation, the Sobol sequence with affine scramble and digital shift will be used~\cite{ref_article4,ref_article5,ref_book5},
which is a standard implementation in Matlab.

\section{Experiment Setup and Benchmarks}

To investigate the possible effect of two different tuning methods on the performance of the FA, two benchmark functions and  a design problem are used in this study. The two benchmark functions are the sphere function and Rosenbrock's banana function. The former is a convex, separable function, whereas the latter is a non-convex, non-separable function. The design problem is a non-convex, nonlinear spring design problem, subject to four constraints.

\subsection{Experimental Setup for FA Parameters}

In the standard FA, there are three parameters $\theta$, $\beta$ and $\gamma$ to be tuned. These parameters of the FA can typically take the following values:

\begin{itemize}
\item Population size: $n = 20$ to $40$ (up to $100$ if necessary).
	
\item $\beta = 0.1$ to $1$ , $\gamma = 0.01$ to $10$, though typically, $\beta=1$ and $\gamma = 0.1$.
	
\item  $\alpha_0 = 1$, $\theta = 0.9$ to $0.99$ (typically, $\theta = 0.97$). where $\alpha = \alpha_0  \theta^{t}$.
	
\item Number of iterations: $t_{\max} = 100$ to $1000$.
\end{itemize}

For simplicity in this study for both MC and QMC, the ranges of these parameter values will be further narrowed down, as shown in Table~\ref{tab1}.

\begin{table}
\begin{center}	
\caption{Experimental setting for MC and QMC Simulations}\label{tab1}
\begin{tabular}{|c|c|c|}
\hline
Initialization Values & Monte Carlo & Quasi-Monte Carlo \\
\hline
Population size &  20 & 20\\
Number of MC/QMC runs &  10 & 10\\
Number of iterations & 1000 & 1000\\ \hline
Parameter ranges of & & \\
$\theta$ & $[0.9,  1.0]$ & $[0.9, 1.0]$ \\
$\beta$ & $[0, 1]$ & $[0, 1]$ \\
$\gamma$  & $[0.5,  2.5]$ & $[0.5,  2.5]$ \\
\hline
\end{tabular}
\end{center}
\end{table}

\subsection{Benchmark Functions}

Optimization algorithms are typically assessed using a diverse set of standard benchmark functions to validate their efficiency and reliability. Researchers evaluate these algorithms by comparing their performance across a wide range of more than two hundred benchmark functions. The choice of benchmarks lacks standardized criteria, but it is essential to use a diverse range of benchmark problems, including different modes, separability,
dimensionality, linearity and nonlinearity. Numerous benchmark collections, including CEC suites, and those referenced in articles, for example, Jamil and Yang~\cite{ref_article6}, are available online.

While test functions are usually unconstrained, real-world benchmark problems originate from various applications with complex constraints and large datasets. This present work will assess the FA's parameter settings using three test benchmarks.

\begin{enumerate}
\item The Rosenbrock function is a nonlinear benchmark~\cite{ref_article7}, which is not convex in the $D$-dimensional space. It is written as
\begin{equation}
f(\x) = (1-x_1)^2+ \sum_{i=1}^{D-1} \left[ 100\left(x_{i+1}-x_i^2\right)^2  \right], \quad \x \in \mathbb{R}^D,
\end{equation}
where
\begin{equation}
-30 \le x_i \le 30, \quad i=1,2,..., D.
\end{equation}
Its global minimum is located at $\x^* = (1, \ldots, 1)$
with $f_{\min}(\x^*) = 0$.

\item The sphere function is a convex benchmark in the form
\begin{equation}
	f(\x) = \sum_{i=1}^D x_{i}^{2}, \quad \x \in \mathbb{R}^D,
\end{equation}
where
\begin{equation}
\quad -10 \leq x_i \leq 10, \quad i=1,2,..., D.
\end{equation}
Its global minimum is located at $\x^* = (0, \ldots, 0)$ with $f(\x^*) = 0$.

\item The spring design is an engineering design benchmark with
three decision variables and four constraints~\cite{ref_article1}.
\begin{equation}
	\textrm{Minimize } \; f(\x) = (2 + x_3) x_1^2 x_2,
\end{equation}
subject to
\begin{align}	
	& g_1(x) =	1 - \frac{x_2^{3}x_3}{71785x_1^{4}} \leq 0, \nonumber \\
	& g_2(x) = \frac{4x_2^2 - x_1x_2}{12566(x_2x_1^{3} - x_1^{4})} + \frac{1}{5108x_1^{2}} - 1 \leq 0, \nonumber \\
	& g_3(x) = 1 - \frac{140.45x_1}{x_2^{2}x_3} \leq 0, \nonumber \\
	& g_4(x) = \frac {x_1+x_2}{1.5} \leq 0. \nonumber
\end{align}
The simple bounds for design variables are
\begin{equation}
0.05 \le x_1 \le 2.0, \quad 0.25 \le x_2 \le 1.3,
\quad 2.0 \le x_3 \le 15.0.
\end{equation}
The best optimal solution found so far in the literature is
\begin{equation}
\x_*=[0.051690, \; 0.356750, \; 11.287126],
\quad f_{\min}(\x_*)=0.012665.
\end{equation}
It is worth pointing out that the constraints in this optimization problem are handled by using the standard penalty method.

\end{enumerate}

\section{Results and Hypothesis testing}

To test the possible effects of different tuning methods on the performance of the FA, a set of 10 runs have been carried out using both MC and QMC methods over three different optimization problems. All problems and runs use the same maximum number of 1000 iterations. Based on the numerical experiments, two hypotheses are proposed and the paired Student's t-tests will be used for comparison.

The two hypotheses to be tested are as follows
\begin{enumerate}
\item[] {\bf Hypothesis H1}: Parameter Tuning methods (MC or QMC) have no significant effect on the fitness values obtained, for a given optimization problem. \\

\item[] {\bf Hypothesis H2}: For a given algorithm, its performance on different problems is not affected by the parameter tuning method used.

\end{enumerate}

\subsection{Testing the First Hypothesis}

For the MC simulation, the parameters of the FA ($\theta$, $\beta$ and $\gamma$) are taken from uniform distributions in the ranges given in Table~\ref{tab1}. Similarly, for the QMC simulation, the parameters are taken from a scrambled Sobol sequence and then mapped into the proper ranges of the parameters. For every objective function, the optimal fitness value obtained along with the corresponding optimal solution, and the parameter values from MC and QMC are recorded for post-processing and hypothesis testing.

\begin{table}[!htbp]
\begin{center}
\caption{Rosenbrock function. \label{table-1}}
	\begin{tabular}{|l|r|r|}
		\hline
	Fitness Values	& Monte Carlo & Quasi-Monte Carlo \\
		\hline
Run 1 & 4.0075e-01  &  2.0416e-01 \\
Run 2 & 3.9088e-01  &  2.7798e-01 \\
Run 3 & 7.2409e-02  &  7.0090e-02 \\
Run 4 & 5.1595e-02  &  3.9830e-01 \\
Run 5 & 3.3661e-01  &  3.2203e-02 \\
Run 6 & 8.3948e-02  &  4.5692e-01 \\
Run 7 & 1.3567e-01  &  7.1474e-02 \\
Run 8 & 5.4517e-02  &  3.7071e-01 \\
Run 9 & 1.1233e-01  &  1.7295e-01 \\
Run 10 & 2.1764e-01  &  9.7198e-01 \\
		\hline
		Mean & $0.1856$ & $0.3027$\\
		\hline
		Std Deviation & $0.1408$ & $0.2776$\\
		\hline
		Test h-value &\multicolumn{2}{|c|}{$0$} \\
		p-value &\multicolumn{2}{|c|}{$0.2498$}\\
		\hline
	\end{tabular}
\end{center}
\end{table}

The t-tests are then used to test the hypotheses. According to the standard t-test criteria, the $h$ and $p$ values obtained from the paired t-tests will determine whether the null hypothesis should be rejected or not. The values for 10 runs obtained from the MC and QMC simulations for each objective function are listed in Table~\ref{table-1} to Table~\ref{table-3}.

\begin{enumerate}
\item {\it Rosenbrock Function.}
From the results for the Rosenbrock function summarized in Table~\ref{table-1}, a paired t-test has been carried out.
As the p-value is much larger than the threshold value 0.05,
the first hypothesis cannot be rejected. That is to say, there is no strong evidence to say that the fitness values obtained are affected by different tuning methods.

\begin{table}[!htbp]
\begin{center}
	\caption{Sphere function. \label{table-2} }
		\begin{tabular}{|l|c|c|}
		\hline
		Fitness Values	& Monte Carlo & Quasi-Monte Carlo \\
		\hline
Run 1 & 3.0614e-003  &  0.0000e+000 \\
Run 2 & 0.0000e+000  &  2.7361e-242 \\
Run 3 & 1.4977e-001  &  1.7983e-003 \\
Run 4 & 3.1305e-002  &  0.0000e+000 \\
Run 5 & 7.4281e-316  &  0.0000e+000 \\
Run 6 &  7.6838e-003  &  3.0621e-139 \\
Run 7 & 1.3688e-003  &  2.3227e-003 \\
Run 8 & 0.0000e+000  &  0.0000e+000 \\
Run 9 & 9.1990e-110  &  0.0000e+000 \\
Run 10 & 0.0000e+000  &  1.0116e-254 \\
		\hline
		Mean & 1.9318e-02 & 4.1211e-04\\
		\hline
		Std Deviation & 4.6842e-02 & 8.7754e-04\\
		\hline
		Test h-value &\multicolumn{2}{|c|}{$0$} \\
		p-value &\multicolumn{2}{|c|}{$0.2181$}\\
		\hline
		
	\end{tabular}
\end{center}
\end{table}	

\item {\it Sphere Function.}
For the sphere function, the paired t-test has been carried out.
As clearly seen in Table~\ref{table-2}, the high p-value means that there are no significant differences in the results obtained by two tuning methods.

\begin{table}[!htbp]
\begin{center}
\caption{Spring design. \label{table-3}}
	\begin{tabular}{|l|r|r|}
	\hline
	Fitness Values	& Monte Carlo & Quasi-Monte Carlo \\
	\hline
	Run 1 & $0.016590$ & $0.030303$ \\
	Run 2	& $0.035127$ & $0.226819$ \\
	Run 3	& $0.033739$ & $0.0171432$ \\
	Run 4	& $0.026841$ & $0.047312$ \\
	Run 5 & $0.119560$ & $0.039970$ \\
	Run 6 & $0.017505$ & $0.021963$ \\
	Run 7	& $0.027859$ & $0.053268$ \\
	Run 8 & $0.021918$ & $0.148673$ \\
	Run 9	& $0.021212$ & $0.0193273$ \\
	Run 10	& $0.021681$   & $0.071922$ \\
	\hline
	Mean & $0.0342$ & $0.0567$\\
	\hline
	Std Deviation & $0.0306$ & $0.0621$\\
	\hline
	Test h-value &\multicolumn{2}{|c|}{$0$} \\
	p-value &\multicolumn{2}{|c|}{$0.3179$}\\
	\hline	
\end{tabular}
\end{center}
\end{table}

\item {\it Spring design problem.}
For a realistic design problem with highly nonlinear constraints,
a similar hypothesis test, based on the results in Table~\ref{table-3}, indicates again no significant differences in
the results between MC and QMC.
\end{enumerate}

\subsection{Testing the Second Hypothesis}

\begin{table}[!htbp]
\begin{center}
\caption{Mean fitness values  for three different problems.
\label{table-4}}
\begin{tabular}{|l|c|c|}
\hline
Problem & Monte Carlo & Quasi-Monte Carlo \\
\hline
Rosenbrock & 0.1856 & 0.3027 \\
Sphere & 0.0193 & 0.0004 \\
Spring Design & 0.0342 & 0.0567 \\
\hline
	Test h-value &\multicolumn{2}{|c|}{$0$} \\
	p-value &\multicolumn{2}{|c|}{$0.7259$}\\
	\hline	
\end{tabular}
\end{center}
\end{table}

The tests of the first hypothesis for all three problems give
a consistent conclusion that there are no significant
differences in the results obtained by the FA whatever the tuning methods were used. To see if this is consistent with the group means, the mean fitness values from three separate problems are tested using the same t-test.

The t-test results are summarized in Table~\ref{table-4}, which again shows that the null hypothesis (H2) holds. That is to say, there is no statistically significant support for one tuning method being better than the other.

This conclusion is a bit surprising from the perspective that the QMC method usually produces better results than the standard MC method for multiple dimensional numerical integrals. This study seems to show that both MC and QMC methods produce similar results for the parameter tuning purpose.

\subsection{F-Test for Variances}
The hypothesis tests so far show that no significant differences were found in the mean objective values obtained by MC and QMC. However, the same level of mean values does not necessarily give the same level of variances. Thus, it is useful to carry out the test of variances. For this purpose, we use the two-sample F-test to see if the variances for the spring design problem obtained by MC and QMC are equal.

The F-test using the same data shown in Table~\ref{table-3} gives $h=1$ and
p-value $p=0.0262$, which is smaller than the critical value 0.05. This means that there is a sufficient difference in variances to reject the null hypothesis. Therefore, it can be concluded that there are no significant differences in mean values obtained by the FA using MC and QMC, but there are some statistically noticeable differences in their corresponding variances
of the objective values.

However, it is worth pointing out that the statistical tests that have been carried out here are mainly to test the differences in means using paired t-tests. In addition, the sample size of 10 is relatively small, thus it may be possible that further more extensive tests may reveal that more comprehensive results may not be  completely consistent with this preliminary conclusion.

\section{Conclusion and Future Work}

From the simulation results of the three different optimization benchmarks, there is not enough evidence to reject the null hypotheses. For hypothesis 1, surprisingly, there was no significant difference in the fitness values obtained via MC and QMC.

For Hypothesis 2, the fitness values obtained by MC and QMC simulations also fail to reject the null hypothesis. For all three benchmark functions, similar orders of fitness values were obtained for both tuning methods. The QMC method does not produce significantly better results, when compared to the
standard MC method.

The preliminary study consists of only a small number of optimization problems, it may be the case that other benchmark problems and other algorithms may not show such robustness. Therefore, a further study is required to determine whether the parameter settings of the FA using these two parameter tuning methods exhibit the same property. Furthermore, some detailed statistical analysis and theoretical analysis will be needed to gain insights into the effect of parameter tuning and its potential link to the convergence behavior observed in these numerical experiments. These will form part of the authors' further research topics.

\end{document}